\documentclass[lettersize,journal]{IEEEtran}
\usepackage{amsmath,amsfonts}
\usepackage{algorithmic}
\usepackage{algorithm}
\usepackage{array}
\usepackage[caption=false,font=normalsize,labelfont=sf,textfont=sf]{subfig}
\usepackage{textcomp}
\usepackage{stfloats}
\usepackage{url}
\usepackage{verbatim}
\usepackage{graphicx}
\usepackage{cite}
\usepackage{multirow}
\usepackage{booktabs}
\usepackage{hyperref}
\usepackage{gensymb}
\begin{document}





\title{\textbf{PinchBot: Long-Horizon Deformable Manipulation with Guided Diffusion Policy}}



\author{Alison Bartsch$^1$, Arvind Car$^1$, Amir Barati Farimani$^1$
\thanks{$^1$A. Bartsch, A. Car, and A. B. Farimani are with the Department of Mechanical Engineering at Carnegie Mellon University, Pittsburgh, PA, 15213, USA (e-mail: {\tt\small abartsch@andrew.cmu.edu}, {\tt\small acar@andrew.cmu.edu}, {\tt\small barati@cmu.edu})}
}


\maketitle

\begin{abstract}
Pottery creation is a complicated art form that requires dexterous, precise and delicate actions to slowly morph a block of clay to a meaningful, and often useful 3D goal shape. In this work, we aim to create a robotic system that can create simple pottery goals with only pinch-based actions. This pinch pottery task allows us to explore the challenges of a highly multi-modal and long-horizon deformable manipulation task. To this end, we present PinchBot, a goal-conditioned diffusion policy model that when combined with pre-trained 3D point cloud embeddings, task progress prediction and collision-constrained action projection, is able to successfully create a variety of simple pottery goals. For experimental videos and access to the demonstration dataset, please visit our project website: \href{https://sites.google.com/andrew.cmu.edu/pinchbot/home}{https://sites.google.com/andrew.cmu.edu/pinchbot/home}.
\end{abstract}


\section{Introduction}

Pottery creation is a complicated art form that requires a precise sequence of dexterous and delicate actions to slowly morph a piece of clay from a semantically meaningless blob to a meaningful and often useful 3D structure. In this work, we are particularly focused on freehand pinch-based pottery in which the artist creates the objects using their hands without tools or a pottery wheel. This pinch-based pottery task is a very difficult and long-horizon deformable manipulation task that allows us to explore and develop robotic systems that can reliably interact with deformable objects. Deformable objects are considered difficult to manipulate due to open challenges of state representation \cite{zhu2024point, chen2024differentiable}, self-occlusion \cite{deng2024general, zhou2025ssfold, tian2025diffusion}, complex state changes from interaction \cite{zhaole2024dexdlo, jamdagni2024robotic}, and the often long-horizon nature of the manipulation tasks themselves. Existing deformable shaping works tend to focus on shorter "long horizon" tasks such as creating alphabet shapes in  clay \cite{shi2024robocraft, shi2023robocook, bartsch2024sculptbot, bartsch2024sculptdiff, bartsch2024llm, zhang2025manipulating}, or rolling, folding and cutting dough \cite{qi2022learning, lin2022planning, bauer2024doughnet, li2023dexdeform, you2025make, ondras2022robotic, kim2022planning}. While these goals are difficult, they typically only require less than 10 actions to create from an initial clay shape. This makes the task much easier for planning frameworks, as there is a relatively short horizon in which the trajectory needs to be optimized over. Similarly, this simplifies the challenges of policy learning as the sequence of actions the policy must learn is much shorter. In this work, we argue that the pinch pottery task is a true long horizon task that will further test the ability of our system to handle these complex long-horizon goals. Additionally, the pinch pottery task does not have an explicit order of actions that are correct. Many combinations of the same actions will create the goal shape, but not all combinations of these actions will create the goal shape. The high multimodality of the trajectories along with the long-horizon nature of the task will allow us to evaluate and benchmark how well diffusion policy model variants \cite{chi2023diffusion, bartsch2024sculptdiff, ze20243d} are able to complete the task.



\begin{figure}
    \centering
    \includegraphics[width=1.0\linewidth]{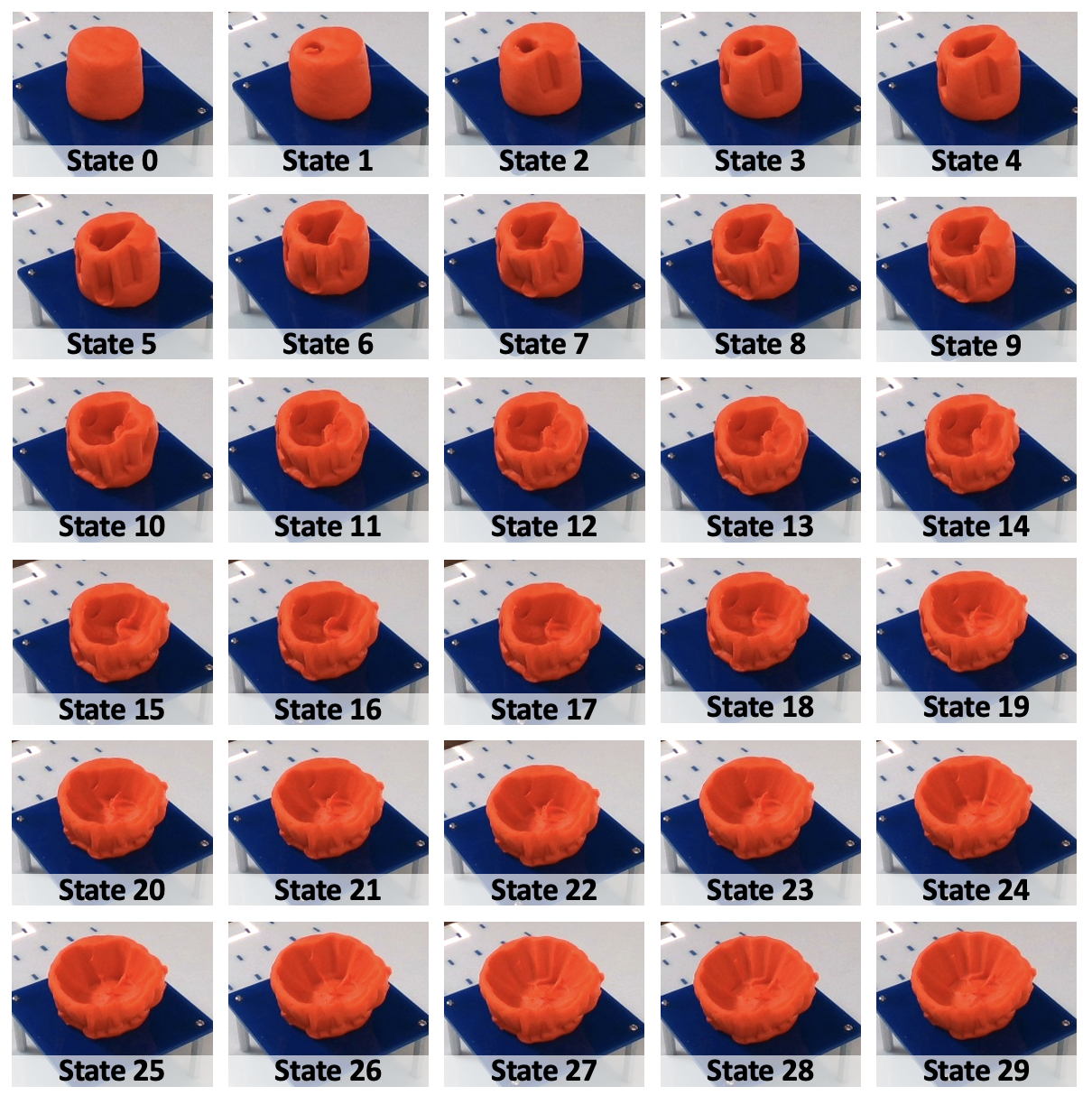}
    \caption{A full pottery sculpting sequence of our trained PinchBot policy.}
    \label{fig:overarching_figure}
\end{figure}

In this work, we present PinchBot a goal-conditioned variant of diffusion policy that employs pre-trained 3D point cloud embedding models, task-progress prediction and collision-constrained projection to successfully create a variety of pottery bowls with only pinch actions. We propose that this pinch pottery task is a true long-horizon deformable manipulation task that allows us to better evaluate learning frameworks. The key contributions of this work are as follows:
\begin{itemize}
    \item We provide a public dataset of human demonstrations collected with kinesthetic teaching on the Franka robot for the pinch pottery task. 
    \item We present a single goal-conditioned policy that is able to adapt the outputted sculpting sequences to successfully create a range of final pottery goals.
    \item We explore how choices in task guidance, pre-training and point cloud embedding model impacts diffusion policy performance for these long-horizon tasks.
    \item We demonstrate how collision-constrained action projection of the policy trajectories is able to improve performance and consistency of the policies. 
\end{itemize}






\section{Related Work}
\textbf{Point Cloud-Based Behavior Cloning:} Across manipulation applications, past work has demonstrated that point clouds improve performance of behavior cloning systems. In \cite{bartsch2024sculptdiff, zhu2024point, ze20243d, yoo2024ropotter}, researchers demonstrate the benefits of point cloud-based diffusion policy \cite{chi2023diffusion}. Additionally, \cite{zhu2024point} demonstrates point cloud state observations improve action chunking transformer \cite{zhao2023learning} across simple manipulation tasks. In \cite{bartsch2024sculptdiff}, researchers incorporate the large transformer model PointBERT \cite{yu2022point} pre-trained on the ShapeNet reconstruction task \cite{chang2015shapenet} with goal-conditioning for the clay shaping task. In \cite{yoo2024ropotter}, researchers leverage a lightweight modified version of the PointNet \cite{qi2017pointnet} model from \cite{ze20243d} for the wheel-based pottery task. While these methods have demonstrated the effectiveness of point cloud conditioned diffusion policy for clay manipulation, in this work, we aim to explore how choice of point cloud model, pre-training and goal-conditioning impact performance on the more challenging task of pinch-based pottery.


\begin{figure*}
    \centering
    \includegraphics[width=0.95\linewidth]{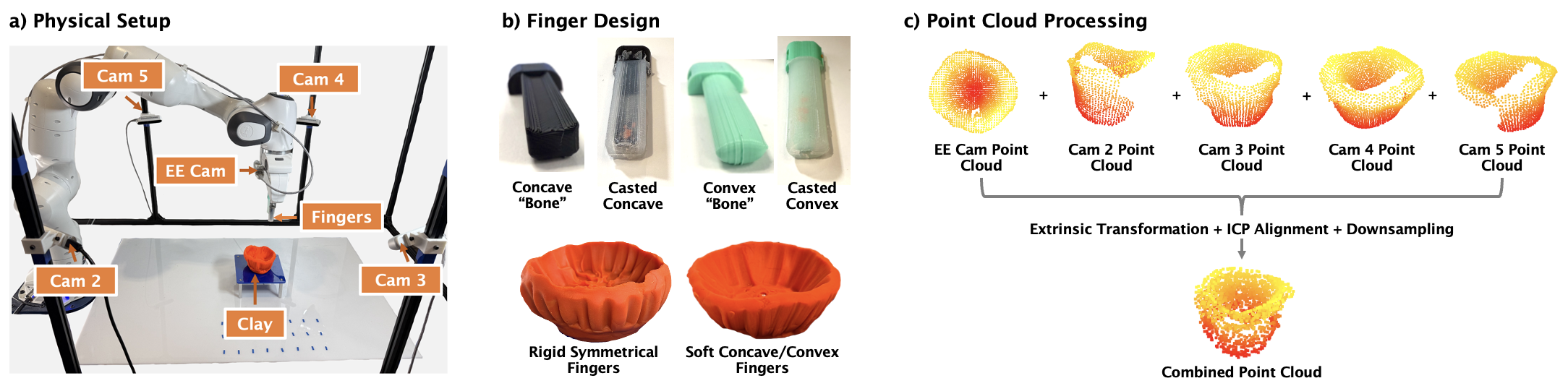}
    \caption{The hardware setup. a) The workspace for our experiments consisting of 4 Intel RealSense D415 cameras, and an elevated stage the clay remains upon. b) The concave and convex finger designs for the pinch pot task. Both fingers consist of a 3D printed concave or convex "bone" casted in EcoFlex for a soft "skin". This design choice helps soften the indentations in the clay creating a smoother surface. c) The point cloud processing pipeline, where the clay is isolated with color thresholding in each camera frame, and then the point clouds are projected into the robot's coordinate frame to be combined.}
    \label{fig:hardware_design}
\end{figure*}

\textbf{Clay Sculpting:} There have been many works that develop robotic systems to shape goals in clay. In \cite{shi2024robocraft}, researchers train a dynamics model to predict deformation dynamics for pinch-based actions. In a follow-up work, they extend the framework to incorporate tools \cite{shi2023robocook}. There have been alternate methods developing dynamics models with pre-trained point cloud embeddings \cite{bartsch2024sculptbot}, or to handle topological state changes \cite{bauer2024doughnet}. Beyond dynamics modeling, imitation learning has shown promise for a set of simple shapes \cite{bartsch2024sculptdiff}. More recently, researchers have explored leveraging large language models for direct action planning \cite{bartsch2024llm}, or for sub-goal generation \cite{bartsch2025planning, you2025make}. While each of these methods explores the very challenging task of clay shaping, they create goal shapes that require approximately 10 or fewer actions. In this work, we argue that evaluating methods with longer-horizon tasks such as pinch pottery could better highlight performance distinctions between methods. In \cite{yoo2024ropotter}, researchers successfully created a wheel-based pottery policy. Although this is a longer-horizon task, because the clay is being rotated, the complexity of the policy itself is much simpler than that of pinch pottery. The state changes the 0DoF end-effector makes will be applied symmetrically to the entire clay surface at that radius and height. Whereas the pinch-pottery task contains more spatial and rotational reasoning challenges.






\section{Pinch-Pot Task}
In this work, we are exploring the challenging long-horizon task of pinch-based pottery. In this section, we will define the task of pinch pottery as well as the action and trajectory representations (Section \ref{sec:task}), show hardware design choices (Section \ref{sec:finger}), describe the full 3D vision point cloud processing system (Section \ref{sec:perception}), and discuss the demonstration data collection process (Section \ref{sec:data}).


\subsection{Task Definition}
\label{sec:task}
We define the pinch pottery task as given a cylinder of clay of varying dimensions initialized in the scene and a 3D point cloud of the goal pottery shape, execute a sequence of pinch actions until the goal pottery shape has been reached. We define each pinch action as the position and orientation of the end-effector, the final distance between the fingertips, and a termination parameter $\gamma$ to identify the final action in each trajectory ($a_{pinch} = [x,y,z,R_x, R_y, R_z, d_{ee}, \gamma]$). In practice, all non-final actions have $\gamma = -1$, and final actions have $\gamma = 1$. For each pinch pottery trajectory, a 3D point cloud is captured before and after each pinch action. The trajectory is thus a sequence of point cloud state and 8D action pairs.





\subsection{Finger Design}
\label{sec:finger}
To ensure the final quality of the pinch pots created by our robotic system, we took care in designing fingertips to reduce the harsh indentation lines that can be common with robotic pinch-based shaping \cite{bartsch2024sculptbot, shi2024robocraft, bartsch2024sculptdiff, bartsch2025planning}. A full visualization of the finger design are shown in Figure \ref{fig:hardware_design}b. The two key aspects of our design are 1) using a concave and convex design to create a curved surface with each squeeze, and 2) casting a 3D printed "bone" with a soft EcoFlex "skin" to reduce harsh indentations. Both the "bone" and "skin" are concave or convex in shape for each finger respectively. The design choice of asymmetrical fingers improves the visual quality of the pottery, but it does widen the rotational action space substantially as the gripper can no longer be considered symmetrical. In practice, this means that we must consider the full $360\degree$ rotation of the $R_z$ component of the action, instead of collapsing the rotational space into a smaller range as was done in previous work \cite{bartsch2024sculptdiff}.

\subsection{Perception}
\label{sec:perception}
We chose 3D point clouds as the state representation for this task as creating pottery is an inherently 3D task, and point clouds have been shown to work well as a 3D representation for diffusion policy \cite{bartsch2024sculptdiff, yoo2024ropotter, chi2023diffusion}. To get a complete view of the pot through all stages of creation, we use 4 cameras fixed in the scene as well as an end-effector mounted camera to provide a top view. This top view is critical to capture the interior of the bowl, particularly in the earlier phases when the opening is very small and is highly occluded from the fixed side-view cameras by the pottery walls. A visualization of the camera setup is shown in Figure \ref{fig:hardware_design}a. To get the final point cloud of the clay, a point cloud is captured by each of the 5 cameras. Next, each point cloud is transformed into the robot coordinate frame. 
For each camera, this requires the extrinsic calibration matrix, and for the end-effector camera this also requires the current end-effector pose. After each point cloud has been transformed, the clay points are isolated through position and color-based thresholding. Finally, the point cloud alignment between the 4 fixed camera views and the end-effector camera is tuned with iterative closest point (ICP) to adjust for variations due to proprioceptive error. The final combined point cloud is then uniformly downsampled to a fixed 2048 points. A visualization of the point cloud processing shown in Figure \ref{fig:hardware_design}c.





\begin{figure*}
    \centering
    \includegraphics[width=0.95\linewidth]{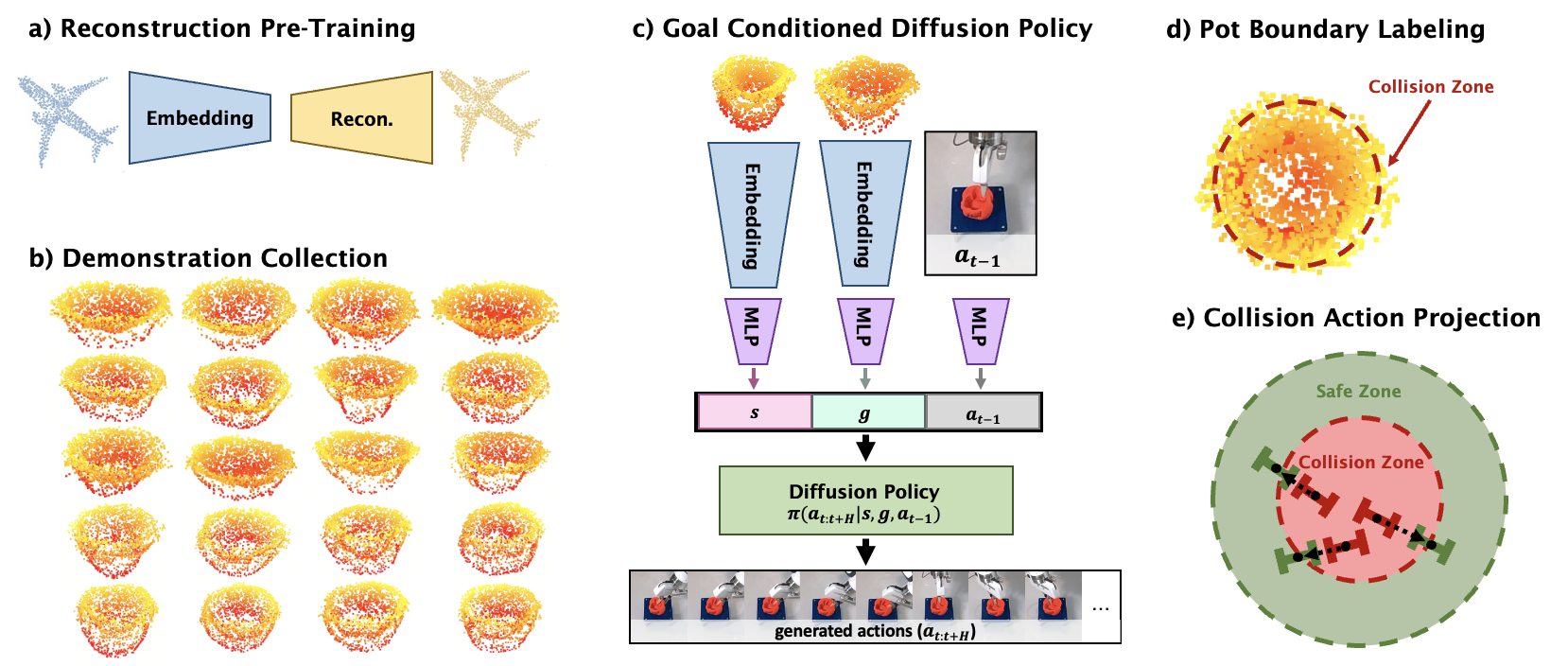}
    \caption{The PinchBot method. a) Pre-training embeddings for the reconstruction task. b) Collecting 20 real-world pinch pottery demonstrations with kinesthetic teaching. c) Training goal-conditioned diffusion policy variants. d) Pottery real-time diameter prediction for collision projection. e) Project all unsafe action positions (i.e. actions that would collide with the existing clay wall) into the safe zone. The safe zone boundary is determined by the current predicted diameter.}
    \label{fig:method}
\end{figure*}

\subsection{Data Collection}
\label{sec:data}
In this work, we collect a demonstration dataset of 20 pinch pottery trajectories with kinesthetic teaching. We vary the initial clay cylinder dimensions between 5 and 8 cm in height (assuming constant clay volume). We vary the final pottery bowl diameter between 7 and 12 cm. Due to the constant clay volume, larger clay diameters are also correlated with angled walls, whereas smaller diameters are correlated with vertical walls. The demonstration trajectories vary in length between 21 and 31 individual pinch actions. Before training, we apply a rotation transformation about the z-axis in increments of $2\degree$ to both the point cloud and the action to increase our dataset size to 3600 trajectories. A visualization of the point clouds captured of the final pottery states for each demonstration is shown in Figure \ref{fig:method}b. For the dataset, see the project website.


\section{Method}
The key components of PinchBot are that of embedding pre-training (Section \ref{sec:pretraining}), goal conditioning (Section \ref{sec:goal_condition}), sub-goal guidance (Section \ref{sec:subgoal}), task progress guidance (Section \ref{sec:task_progress}), and collision action projection (Section \ref{sec:colaction}). A visualization of the framework is shown in Figure \ref{fig:method}.

\subsection{Point Cloud Embedding Pre-Training}
\label{sec:pretraining}
We use pre-training as a method of overcoming the challenges of training our point cloud embedding models with a very small dataset of real-world demonstrations. In this work, we are comparing the performance of two point cloud embedding models that have been demonstrated to work well for point cloud-based diffusion policy in the literature, PointBERT \cite{yu2022point}, and a modified version of PointNet from \cite{ze20243d} (referred to as DP3 PointNet in this paper). In \cite{ze20243d} and \cite{yoo2024ropotter}, researches directly use the DP3 PointNet embedding model without any pre-training for deformable shaping tasks. However, in previous work, we found that pre-training the PointBERT model works well for goal-conditioned point cloud diffusion policy \cite{bartsch2024sculptdiff}. With this work, we aim to explore the behavior differences between goal-conditioned policies trained with each point cloud embedding model. We pre-train both the PointBERT and DP3 PointNet models on the ShapeNet \cite{chang2015shapenet} reconstruction task before fine-tuning the models on the diffusion policy objective. For fine-tuning each embedding model, we add a 3-layer MLP head to project the point cloud embedding into the low dimensional space of size 512. 







\subsection{Goal Conditioning}
\label{sec:goal_condition}
In this work we aim to train a single pinch pottery policy that can adapt to multiple pottery diameter goals. Goal conditioning is essential to having an adaptable and controllable policy. Additionally, goal conditioning combined with termination prediction allows the policy to be fully autonomous without any thresholding to determine stopping conditions. Without goal conditioning, a single policy must be trained for a single goal, which simplifies the problem \cite{yoo2024ropotter}. For goal conditioning, the goal point cloud is embedded with the point cloud model of choice, and that latent goal embedding is concatenated with the state embedding and previous action to condition the iterative denoising of the action trajectory (shown in Figure \ref{fig:method}c).






\subsection{Subgoal Diffusion Policy}
\label{sec:subgoal}
Instead of conditioning the denoising process by the concatenated latent state, goal and previous action vectors, for our version of sub-goal diffusion policy, the policy is conditioned by the concatenated latent state, $N$ latent sub-goals and the previous action vector. The parameter $N$ is determined based on the action prediction horizon and the sub-goal step size, i.e. if the prediction horizon is 16 actions and the sub-goal step size is 8, then $N = 2$. By providing intermediate sub-goals we are exploring if this intermediate state guidance helps the policy with the long-horizon action generation.

\begin{table*}[]
\caption{\textbf{Pinch Pottery Quantitative Results}. Three runs were conducted per embedding/training objective variant.}\label{tab:results} 
\centering
\scriptsize
\begin{tabular}{@{\extracolsep{\fill}}llllllllllll}
    \toprule
    & & \multicolumn{3}{l}{\textbf{---------------- 8cm Diameter ----------------}}  & \multicolumn{3}{l}{\textbf{--------------- 10cm Diameter ---------------}} & \multicolumn{3}{l}{\textbf{--------------- 12cm Diameter ---------------}} \\
    \midrule
    & & CD [mm]   & EMD [mm] & MSE [mm] & CD [mm]  & EMD [mm] & MSE [mm]  & CD [mm]   & EMD [mm]  & MSE [mm] \\
    \hline
    \midrule
    
    \centering \multirow{3}{*}{\textbf{PointBERT}}   & Binary Pred. & 9.1 $\pm$ 0.5 & 7.8 $\pm$ 0.6 & 0.02 $\pm$ 0.02 & \textbf{7.1 $\pm$ 0.2 }& \textbf{6.3 $\pm$ 0.5} & \textbf{0.05 $\pm$ 0.02} & \textbf{7.9 $\pm$ 0.0} & \textbf{7.2 $\pm$ 0.2} & \textbf{0.87 $\pm$ 0.18} \\
    & Cont. Guid. & 9.9 $\pm$ 0.5 & 9.2 $\pm$ 0.5 & \textbf{0.02 $\pm$ 0.01} & 7.3 $\pm$ 0.2 & 6.5 $\pm$ 0.4 & 0.36 $\pm$ 0.06 & 8.2 $\pm$ 0.3 & 7.2 $\pm$ 0.4 & 1.29 $\pm$ 0.68 \\
    & Sub-Goal & \textbf{8.2 $\pm$ 0.3} & \textbf{6.9 $\pm$ 0.5} & 0.04 $\pm$ 0.04 & 9.5 $\pm$ 0.6 & 8.7 $\pm$ 1.0 & 0.73 $\pm$ 0.07 & 9.9 $\pm$ 0.8 & 9.5 $\pm$ 0.6 & 3.48 $\pm$ 0.62 \\
    \midrule

    \centering \multirow{3}{*}{\textbf{DP3 PointNet}}   & Binary Pred. & 9.2 $\pm$ 0.6 & 8.6 $\pm$ 0.7 & 0.18 $\pm$ 0.09 & 7.2 $\pm$ 0.3 & \textbf{6.5 $\pm$ 0.4} & \textbf{0.10 $\pm$ 0.10} & 9.9 $\pm$ 0.4 & 9.3 $\pm$ 0.7 & 1.38 $\pm$ 0.54 \\
    & Cont. Guid. & \textbf{8.5 $\pm$ 0.9} & 8.6 $\pm$ 1.1 & 0.62 $\pm$ 0.48 & \textbf{7.0 $\pm$ 0.2} & \textbf{6.5 $\pm$ 0.4} & 0.39 $\pm$ 0.13 & \textbf{9.2 $\pm$ 0.2} & \textbf{8.7 $\pm$ 0.1} & \textbf{1.32 $\pm$ 0.11} \\
    & Sub-Goal & 9.1 $\pm$ 1.1 & \textbf{7.9 $\pm$ 0.8} & \textbf{0.06 $\pm$ 0.05} & 9.6 $\pm$ 0.3 & 9.1 $\pm$ 0.5 & 1.11 $\pm$ 0.31 & 10.2 $\pm$ 0.3 & 9.9 $\pm$ 0.6 & 3.88 $\pm$ 0.85 \\
    \bottomrule
\end{tabular}
\end{table*}

\subsection{Task Progress Guided Diffusion Policy}
\label{sec:task_progress}
To continue the exploration of how to best guide our policy for the long-horizon task of pinch pottery, we propose a very simple strategy to have the model predict where along the trajectory the generated action is. We call this task progress guided diffusion policy, as the model is learning to predict how far into the specific demonstration it is. To do so, we modify $\gamma$ from the action to predict a continuous value between $-1$ and $1$ to represent the percentage along the trajectory, where $-1$ represents the first state and $1$ represents the final state. 

\subsection{Collision Action Projection}
\label{sec:colaction}
For our task, slight errors in position and rotation of a single pinch can be catastrophic and result in irrecoverable states, such as creating a hole in the pottery wall. In this work, we develop a collision detection and action projection framework to minimize harmful actions. The process is visualized in Figure \ref{fig:method}d-e. We fit a circle to the current state point cloud by finding the diameter that fits $>95\%$ of points projected into the x,y plane within the circle. We then take the action generated from our policy and identify if the current x,y position lies within the circle. If so, the action position is projected onto the edge of the circle. This modifies the x,y parameters of the action while preserving the z, rotations and fingertip distance components.


\begin{figure*}
    \centering
    \includegraphics[width=1.0\linewidth]{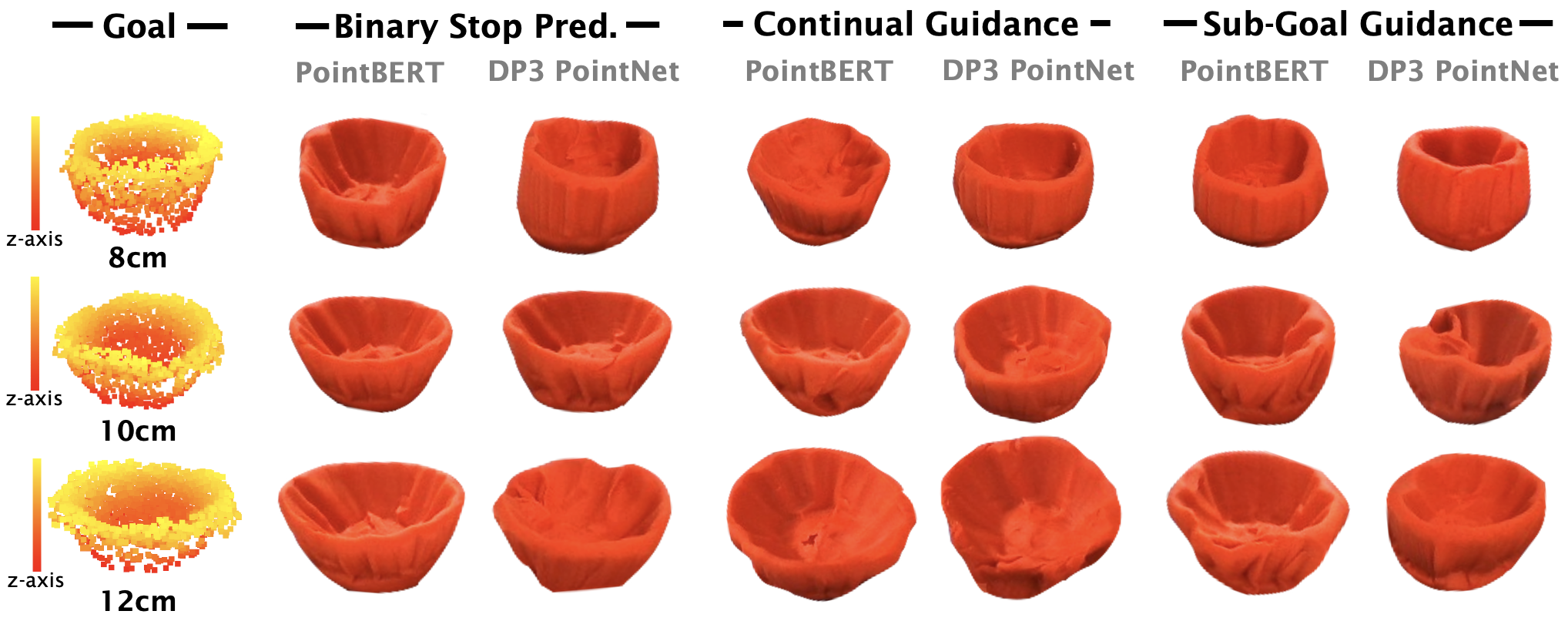}
    \caption{The final shapes conditioned on the 8cm, 10cm and 12cm bowl diameter goals across point cloud embedding and policy variant.}
    \label{fig:guidance_results}
\end{figure*}

\begin{figure*}
    \centering
    \includegraphics[width=1.0\linewidth]{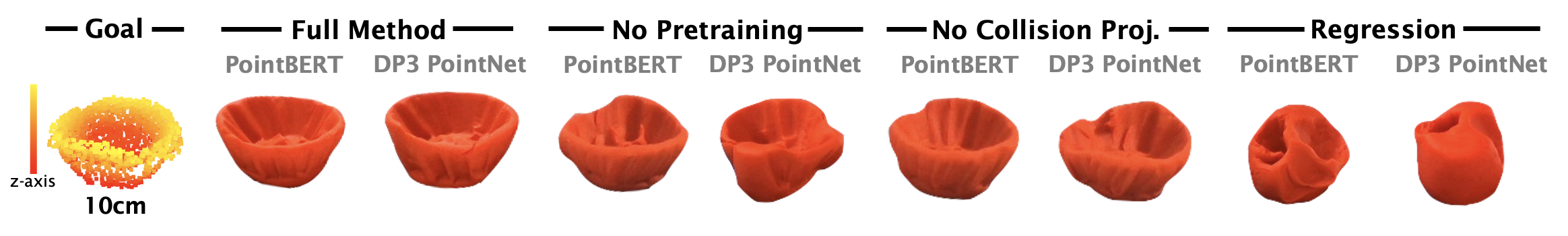}
    \caption{The final shapes for the ablation study variants exploring how pre-training, collision projection and diffusion policy-style training impact the final method performance. }
    \label{fig:ablation_results}
\end{figure*}

\section{Experiments and Results}
Through our experiments, we explore how choices of embedding model (Section \ref{sec:embeddings}), and task progress or sub-goal guidance (Section \ref{sec:progress}) impacts policy behavior. Additionally we conduct an ablation study (Section \ref{sec:ablation}), and analyze the latent space of the point cloud embeddings with T-SNE (Section \ref{sec:tsne}). For each experiment variant, we perform three real-world runs of the policy and present the mean and standard deviation for each evaluation metric. Each variant of diffusion policy was trained with an action prediction horizon of 16, and at test time had an execution horizon of 4 actions before the policy replanned. The sub-goal diffusion policy variants were trained with sub-goals of a step size of 8.






\subsection{Evaluation Metrics}

To evaluate how well each policy is able to create a variety of pottery goals, we evaluate the Chamfer Distance (CD) \cite{rubner2000earth} and Earth Mover's Distance (EMD) \cite{fan2017point} between the final state and goal point clouds for the goal pottery diameters of 8cm, 10cm and 12cm. In addition to these point-wise similarity metrics, we present the mean squared error (MSE) between the goal pottery diameter and the final pottery diameter created by the policy.

\subsection{Impact of Point Cloud Embeddings}
\label{sec:embeddings}
Through our experiments, we find that the PointBERT policies outperform the DP3 PointNet variants across all quantitative metrics (Shown in Table \ref{tab:results}). Additionally, the PointBERT policies require much fewer actions to reach the final bowls (\ref{tab:nactions}). The DP3 PointNet policies take approximately two times more actions than the underlying demonstration trajectories required. The PointBERT policies are able to better determine when to stop and where in the trajectory the state is. Qualitatively, we found the DP3 PointNet policies would struggle to identify when to switch between regions of the pot, and execute additional and unnecessary pinches when the region was already smooth. 



\subsection{Impact of Sub-Goals and Task Guidance}
\label{sec:progress}
We find that task progress guidance improves the DP3 PointNet policy behavior and adaptability to multiple bowl diameters (shown in Table \ref{tab:results}). Task progress guidance has minimal impact on the PointBERT policy's quantitative performance (both the binary prediction and task progress guidance results are within the standard deviation ranges of each other). However, qualitatively the task progress guidance training does improve the PointBERT policy's ability to distinguish $R_x$ end-effector rotation variations depending on the goal point cloud. The $R_x$ of the action directly informs the angle of the pottery wall. For the 8cm diameter bowl, the underlying demonstrations had very small x-axis rotations to create a near vertical wall. With task-progress guidance, the PointBERT policy is able to better distinguish behavior patterns between goals and correctly vary the $R_x$ more widely. In contrast, the PointBERT policy trained with binary prediction is able to successfully match the underlying goal diameter and pottery structure more by accurate positional variations in each squeeze, with some smaller variations in rotation. While both policies are able to adequately meet the goal variations, we believe that the task-progress guidance produces a more expressive and variable policy. On the other hand, the sub-goal guided diffusion policy struggles with the largest bowl diameter of 12cm. We hypothesize that this could be because the demonstration trajectories for the larger diameter bowls have intermediate states of the smaller diameter bowls. With sub-goal conditioning only (i.e. the policy is not provided the final pottery goal for all states), there is a much larger distribution of training data points for creating the smaller diameter bowls.

\begin{table}[]
    \caption{\textbf{Mean Actions for Each Method.} Across goal diameters.}\label{tab:nactions} 
    \centering
    \begin{tabular}{@{\extracolsep{\fill}}llllll}
            \toprule
             \textbf{ } & \textbf{ } & \textbf{\# Actions} \\ 
            \midrule
            \hline
            \multirow{3}{*}{\textbf{PointBERT}} & Binary Pred. & 28.4 $\pm$ 2.6 \\
            & Cont. Guid. & 33.3 $\pm$ 10.7 \\
            & Sub-Goal & 34.8 $\pm$ 5.3 \\
            \midrule
            \multirow{3}{*}{\textbf{DP3 PointNet}} & Binary Pred. & 51.3 $\pm$ 18.5 \\
            & Cont. Guid. & 57.4 $\pm$ 17.5 \\
            & Sub-Goal & 35.4 $\pm$ 4.2 \\
            \bottomrule
            
        \end{tabular}
    \end{table}

\begin{table}[]
\caption{\textbf{Ablation Study}. Binary prediction models on the 10cm goal.}\label{tab:ablation} 
\centering
\scriptsize
\begin{tabular}{@{\extracolsep{\fill}}llllll}
    \toprule
    & & CD [mm]   & EMD [mm] & MSE [mm]  \\
    \hline
    \midrule
    
    \centering \multirow{4}{*}{\textbf{PointBERT}}   & Pretraining &  \textbf{7.1 $\pm$ 0.2} & \textbf{6.3 $\pm$ 0.5} & \textbf{0.05 $\pm$ 0.02}  \\
    & No pretraining & 7.4 $\pm$ 0.2 & 6.6 $\pm$ 0.3 & 0.78 $\pm$ 0.05  \\
    & No col. proj.  & 7.7 $\pm$ 0.3 & 7.9 $\pm$ 0.1 & 1.02 $\pm$ 0.25 \\
    & Regression & 18.9 $\pm$ 1.9 & 17.7 $\pm$ 1.7 & 7.75 $\pm$ 2.26 \\
    \midrule

    \centering \multirow{4}{*}{\textbf{DP3 PointNet}}   & Pretraining & \textbf{7.2 $\pm$ 0.3} & \textbf{6.5 $\pm$ 0.4} & \textbf{0.10 $\pm$ 0.10} \\
    & No pretraining & 8.3 $\pm$ 0.5 & 7.7 $\pm$ 0.7 & 0.80 $\pm$ 0.91  \\
    & No col. proj. & 7.5 $\pm$ 0.1 & 7.1 $\pm$ 0.2 & 0.64 $\pm$ 0.09  \\
    & Regression & 20.7 $\pm$ 0.4 & 20.9 $\pm$ 0.1 & 9.86 $\pm$ 0.28 \\
    \bottomrule
\end{tabular}
\end{table}

\subsection{Ablation Study}
\label{sec:ablation}
Through our ablation study, we explore how pre-training, collision projection and the diffusion training process impact the performance of our method. To do so, we evaluate both PointBERT and DP3 PointNet binary prediction policies on the 10cm diameter goals as compared to the variants with each component removed (quantitative results in Table \ref{tab:ablation}, with qualitative results visualized in Figure \ref{fig:ablation_results}). Through our ablation study we find that pre-training, collision projection and the diffusion policy-styled training itself all contribute to the success of our proposed method. Diffusion policy itself is incredibly important for this method, as our data is highly multimodal. Furthermore, pre-training the point cloud embedding improves the ability of the policies to correctly identify when the goal has been reached.

\begin{figure*}
    \centering
    \includegraphics[width=1.0\linewidth]{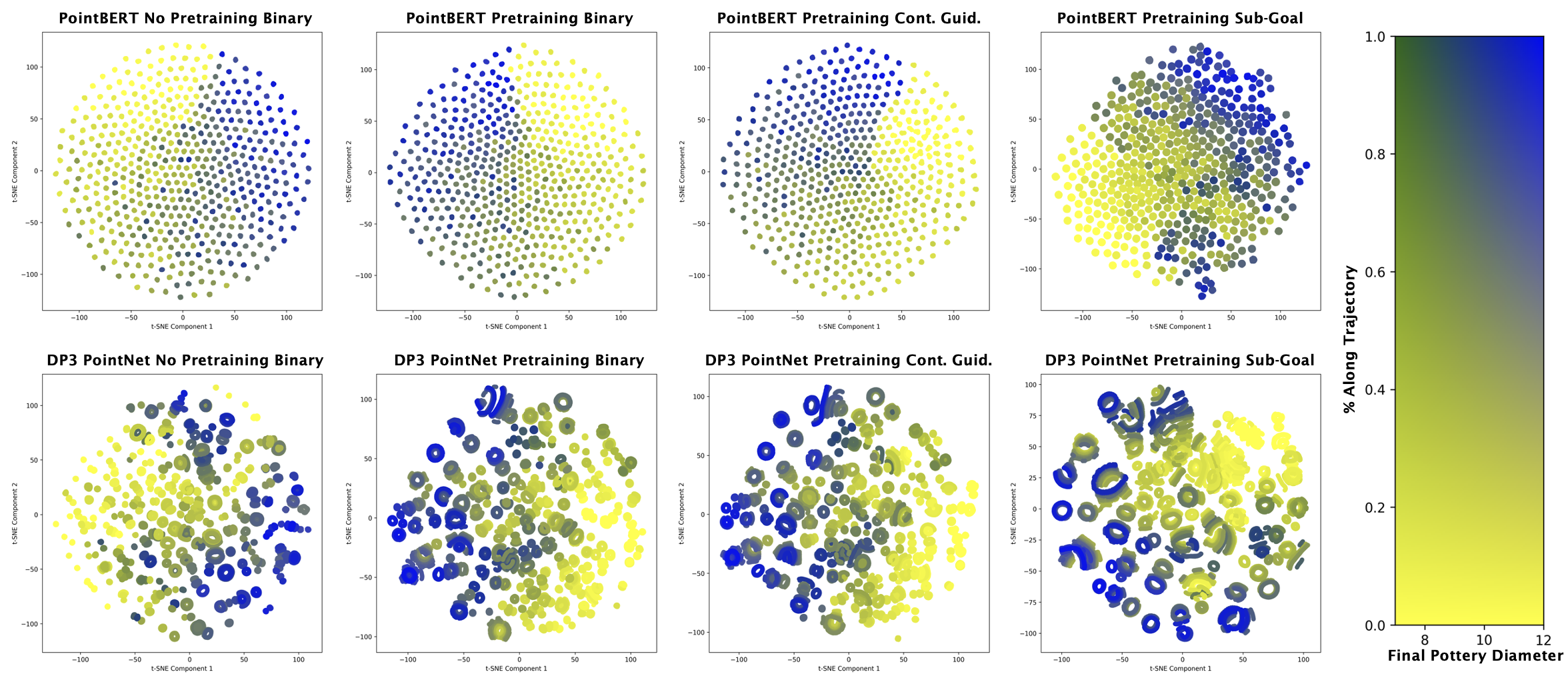}
    \caption{T-SNE plots of the embeddings of the rotation augmented demonstration trajectories across model variations. The colorbar variation along the x-axis corresponds with the final pottery diameter of each trajectory visualized, wheras the y-axis color variation visualizes the point along each respective demonstration trajectory.}
    \label{fig:tsne_embeddings}
\end{figure*}

\subsection{Analysis of Latent Space Visualization}
\label{sec:tsne}
To further investigate performance differences between embedding models and policy variations, we visualize the latent emebeddings of the demonstration pottery trajectories with T-SNE in Figure \ref{fig:tsne_embeddings}. Comparing the visualizations for the two different embedding models, we can see a much more defined distinction along the trajectories in the PointBERT latent space than that of DP3 PointNet. This corroborates the result that the DP3 PointNet model struggles to identify when regions of the pot have been completed and/or the goal has been reached. Furthermore, the PointBERT sub-goal guided policy's latent space has a much worse separation along trajectory progress and final pottery diameters as compared to the other PointBERT policy variants, aligning with the poor performance of the sub-goal policy. Finally, we can see that the DP3 PointNet policy with continuous guidance has the most clear separation in terms of both final pottery diameter and trajectory progress among the DP3 PointNet variations, which aligns with the quantitative results.








\section{Conclusion and Limitations}
In this work, we present PinchBot, an imitation learning framework leveraging pre-trained point cloud embeddings, task-progress guidance and diffusion policy to successfully learn a single, goal-conditioned policy that can adapt sculpting trajectories to create a variety of long-horizon pottery goals. While we have shown the ability for our single policy to have modest adaptation capabilities to a variety of goals with a small quantity of real-world demonstrations, this behavior cloning framework is still somewhat limited in terms of generalizability. Future work could explore creating a subtantially larger dataset, or incorporating dynamics prediction into the policy learning framework with the aim of designing a more widely generalizable and adaptable robotic pottery framework. Alternatively, recent work has explored steering existing diffusion policies with latent space reinforcement learning \cite{wagenmaker2025steering}, which could be applied directly to PinchBot to quickly adapt the policy to a wider range of goals. Regardless of these limitations of generalizability, we present a framework that can successfully learn an adaptable, goal-conditioned pinch pottery policy that is able to produce good quality pottery bowls. The true long-horizon nature of this pinch pottery task allows us to explore policy behavior differences due to choice of emebedding model and pre-training. We hope that in the future, deformable shaping works will benchmark on complicated tasks beyond alphabet shaping goals, as when deployed on these more challenging long-horizon tasks we can better identify nuanced performance differences among methods.


\bibliographystyle{IEEEtran}
\bibliography{references}

\end{document}